\documentclass[12pt, a4paper]{article}
\pdfoutput=1

\usepackage{tgtermes}

\usepackage{amsmath,empheq,amsfonts,bbm}
\usepackage{epic,eepic,epsfig,longtable}
\usepackage{multirow,verbatim}
\usepackage{subfigure}
\usepackage{array}
\usepackage{wasysym}
\usepackage[ruled,linesnumbered]{algorithm2e}

\usepackage[colorlinks,citecolor=blue,linkcolor=blue]{hyperref}
\usepackage{epsfig}
\usepackage{pdflscape}
\usepackage{setspace}
\usepackage{mathrsfs}
\usepackage[figuresleft]{rotating}
\usepackage{amsbsy}
\usepackage{eqnarray}
\usepackage{booktabs}
\usepackage{epstopdf}
\usepackage{color}
\usepackage{cases}
\usepackage{appendix}
\usepackage{lineno}

\makeatletter
\def\singlespace{\def\baselinestretch{1}\@normalsize}

\makeatletter
\def\singlespace{\def\baselinestretch{1}\@normalsize}

\allowdisplaybreaks
\usepackage{verbatim}
\renewcommand{\baselinestretch}{1}
\numberwithin{equation}{section}
\newtheorem{theorem}{Theorem}[section]
\newtheorem{lemma}[theorem]{Lemma}
\newtheorem{definition}{Definition}
\newtheorem{remark}{Remark}
\newcounter{CondCounter}
\newenvironment{proof}{{\noindent\it Proof.}\ }{\hfill $\square$\par}

\begin{document}

\title{\textbf{$L_1$-norm Regularized Indefinite Kernel Logistic Regression}}

\author{
Shaoxin Wang \footnotemark[2] \footnotemark[3]
\and
Hanjing Yao \footnotemark[2] \footnotemark[4]
}
\footnotetext[2]{These authors contributed equally to this work.}
\footnotetext[3]{Corresponding author. School of Statistics and Data Science, Qufu Normal University,  Qufu 273165, China (shxwang@qfnu.edu.cn or shxwangmy@163.com).}
\footnotetext[4]{Dongfang College, ShanDong University of Finance and Economics, Taian 271000, China (yaohanjing1999@163.com).}

\date{\small Last modified on \today}
\maketitle


\begin{abstract}
Kernel logistic regression (KLR) is a powerful classification method widely applied across diverse domains.
In many real-world scenarios, indefinite kernels capture more domain-specific structural information than positive definite kernels.
This paper proposes a novel $L_1$-norm regularized indefinite kernel logistic regression (RIKLR) model,
which extends the existing IKLR framework by introducing sparsity via an $L_1$-norm penalty.
The introduction of this regularization enhances interpretability and generalization while introducing nonsmoothness
and nonconvexity into the optimization landscape.
To address these challenges, a theoretically grounded and computationally efficient proximal linearized algorithm is developed.
Experimental results on multiple benchmark datasets demonstrate the superior performance of the proposed method
in terms of both accuracy and sparsity.
\end{abstract}


\section{Introduction}
\label{sec:introduction}
\label{intro}
Kernel methods represent a fundamental class of machine learning techniques and have gained widespread adoption across diverse domains \cite{SchS02}, including computer vision \cite{Lamp09, GTMLKY16}, natural language processing (NLP) \cite{WSLZH16, Beck14}, and bioinformatics \cite{SVUA04}, among others. The core idea underlying kernel methods is to employ a kernel function that implicitly maps the input data into a high-dimensional feature space, thereby enabling the use of linear models to solve nonlinear learning tasks in the original space. Consequently, the selection of an appropriate kernel function is critical to the performance of the method.
Traditional kernel methods predominantly rely on positive definite (PD) kernels, such as the polynomial kernel and the Gaussian kernel. According to Mercer's Theorem, a PD kernel ensures that the resulting kernel matrix is positive semidefinite (PSD), thereby facilitating the analysis of the learning problem within the framework of reproducing kernel Hilbert spaces (RKHS) \cite{EvPP00}. The PSD property guarantees that the corresponding optimization problem is convex and thus tractable.

However, in many real-world applications, similarity measures or kernel functions may be indefinite. Such indefinite kernels often capture domain-specific structures more effectively but lead to non-PSD kernel matrices, violating Mercer's condition \cite{SchT15}. To address learning with indefinite kernels, Ong et al. \cite{OMCS04} introduced the framework of reproducing kernel Krein spaces (RKKSs), which extends SVM analysis to indefinite settings. In contrast to RKHS, the inner product in RKKS can be indefinite, resulting in a nonconvex optimization problem that is generally more challenging to solve. For instance, while SVM with a PD kernel can be efficiently optimized via quadratic programming, the use of an indefinite kernel may lead to a nonconvex objective function that lacks a global minimizer \cite{OMCS04}.

Kernel logistic regression (KLR) \cite{ZhuH05} is a powerful classification method that has been successfully applied across various domains. For instance, it has been used to accurately classify cancer patients based on genetic profiles \cite{KSKL06} and in speech recognition tasks \cite{KPSV07}. Li et al. \cite{LMPBB13} proposed a generalized composite kernel machine framework based on multinomial logistic regression, which achieved outstanding classification performance on hyperspectral images. Recently, sparse versions of KLR have attracted considerable attention. Katz et al. \cite{KSAKW06} introduced a sparse kernel logistic regression (SKLR) model for speaker identification. Lee et al. \cite{LLAN06} developed an $L_1$-norm regularized KLR (RKLR) and employed a quadratic approximation technique to design an efficient algorithm suitable for large-scale datasets. Another approach, the $L_{1/2}$-norm RKLR, uses an iterative thresholding algorithm to achieve effective feature selection \cite{XPJ13}.

The indefinite kernel logistic regression (IKLR) \cite{LHGYS18} extends traditional KLR by allowing the use of indefinite kernels, thereby introducing nonconvexity into the problem. By leveraging eigenvalue decomposition of the kernel matrix, Liu et al. \cite{LHGYS18} showed that the nonconvex objective function of IKLR can be expressed as a difference of two convex (DC) functions. They designed a concave-convex procedure (CCCP) to solve the optimization problem and further proposed a concave-inexact-convex procedure (CCICP) along with its stochastic variant to improve computational efficiency.

Sparsity plays a vital role in kernel learning, contributing to model efficiency, interpretability, and generalization capability. It aids in feature selection by identifying the most relevant variables and mitigating the influence of irrelevant or noisy features. For example, the $L_1$-norm SVM is widely used for simultaneous learning and feature selection \cite{ZRTH03, BraM98, FuM04, HilK08}. As shown in \cite{ZRTH03}, $L_1$-norm SVM particularly excels in scenarios involving redundant or noisy features. Xue et al. \cite{XSX20} introduced a multiple indefinite kernel feature selection method based on the primal SVM framework with indefinite kernels, which automatically selects relevant features via an $L_1$-norm penalty on kernel combination coefficients. Alabdulmohsin et al. \cite{ACGZ16} demonstrated that $L_1$-norm regularized SVM effectively handles indefinite similarities, achieving high classification accuracy while preserving model sparsity.

In this paper, we propose an $L_1$-norm regularized indefinite kernel logistic regression (RIKLR) model. Although IKLR has proven effective in various classification tasks, it often yields non-sparse solutions that may include noisy features. To address this, we incorporate an $L_1$-norm penalty into the IKLR objective function to promote sparsity. Through eigenvalue decomposition of the indefinite kernel matrix, we show that the resulting objective function can also be decomposed into a DC form. However, the $L_1$-norm regularizer also leads to a non-smooth optimization landscape, precluding the direct application of methods from \cite{LHGYS18}. To overcome this challenge, we develop a novel proximal linearized algorithm tailored for non-smooth DC optimization problems. We also analyze the convergence behavior of the proposed algorithm and evaluate its performance on multiple datasets.

The remainder of this paper is organized as follows. Section \ref{sec:review} provides a brief overview of related work on KLR and IKLR. In Section \ref{sec:PIKLR}, we introduce the $L_1$-norm RIKLR model, describe the proposed algorithm, and present its convergence analysis. Numerical experiments demonstrating the efficacy of our method are detailed in Section \ref{sec:experiments}. Finally, concluding remarks are given in Section \ref{sec:conclusion}. Throughout this paper, vectors are denoted by bold lowercase letters. For a vector $\boldsymbol{\alpha} \in \mathbb{R}^n$, $\boldsymbol{\alpha}^{\top}$ represents its transpose. The $L_1$-norm and Euclidean norm are defined as $\|\boldsymbol{\alpha}\|_1 = \sum_{i=1}^n |\alpha_i|$ and $\|\boldsymbol{\alpha}\| = \sqrt{\boldsymbol{\alpha}^{\top}\boldsymbol{\alpha}}$, respectively. For functions $f, g$ in a function space $\mathcal{H}$, $\langle f, g \rangle_{\mathcal{H}}$ denotes their inner product in $\mathcal{H}$.

\section{A brief review of related KLR models}\label{sec:review}

To clearly situate our work within the existing literature,  in this section we present a concise overview of the most relevant and widely-used KLR models for binary classification tasks.

\subsection{The KLR model}
Let ${(\boldsymbol{x}_i, y_i)}^n_{i=1}$ denote a set of $n$ training samples, where each label $y_i$ belongs to $\{0, +1\}$. The KLR problem is formulated as follows
\begin{equation}
\label{eq.KLR1}
\min_{f \in \mathcal{H}_{\mathcal{K}_+}} \frac{1}{n} \sum_{i=1}^{n} \ln\left(1 + e^{-y_i f(\boldsymbol{x}_i)}\right) + \frac{\lambda}{2} \| f \|_{\mathcal{H}_{\mathcal{K}_+}}^2,
\end{equation}
where $\lambda$ is the regularization parameter, $\mathcal{H}_{\mathcal{K}_+}$ is the reproducing kernel Hilbert space (RKHS) induced by a positive definite kernel $\mathcal{K}_+(\cdot, \cdot)$, and $\| f \|_{\mathcal{H}_{\mathcal{K}_+}}^2 = \langle f, f \rangle_{\mathcal{H}_{\mathcal{K}_+}}$. According to the Representer Theorem \cite{SchHS01}, the minimizer of \eqref{eq.KLR1} yields a discriminant function $f(\cdot)$ of the form
\begin{equation}
\label{eq.optf}
f(\cdot) = \sum_{i=1}^{n} \alpha_i \mathcal{K}(\boldsymbol{x}_i, \cdot),
\end{equation}
where $\boldsymbol{\alpha} = [\alpha_1, \alpha_2, \cdots, \alpha_n]^\top$ is the coefficient vector. Note that we omit the constant term in $f(\cdot)$, and the classification rule is given by $\hat{y} = \mathrm{sign}(f(\boldsymbol{x}))$.

Substituting \eqref{eq.optf} into \eqref{eq.KLR1} leads to the following finite-dimensional optimization problem
\begin{equation}
\label{eq.KLR2}
\min_{\boldsymbol{\alpha} \in \mathbb{R}^n} \frac{1}{n} \sum_{i=1}^{n} \ln\left(1 + \exp\left(-y_i \sum_{j=1}^{n} \alpha_j \boldsymbol{K}_{ij}\right)\right) + \frac{\lambda}{2} \sum_{i,j=1}^{n} \alpha_i \alpha_j \boldsymbol{K}_{ij},
\end{equation}
where $\boldsymbol{K}_{ij} = \mathcal{K}(\boldsymbol{x}_i, \boldsymbol{x}_j)$ denotes the $(i,j)$-th entry of the kernel matrix $\boldsymbol{K}$. Equivalently, \eqref{eq.KLR2} can be expressed in matrix form as
\begin{equation}
\label{eq.KLR3}
\min_{\boldsymbol{\alpha} \in \mathbb{R}^n} \frac{1}{n} \boldsymbol{1}^\top \ln\left(\boldsymbol{1} + \exp(-\boldsymbol{y} \odot (\boldsymbol{K} \boldsymbol{\alpha}))\right) + \frac{\lambda}{2} \boldsymbol{\alpha}^\top \boldsymbol{K} \boldsymbol{\alpha},
\end{equation}
where $\boldsymbol{1}$ is the all-ones vector, $\boldsymbol{y}$ is the label vector, and $\odot$ denotes the Hadamard product.
It is worth noting that the solution to the standard KLR problem is generally non-sparse, which limits its interpretability and compressibility. To promote sparsity in the coefficient vector $\boldsymbol{\alpha}$, one may adopt iterative selection procedures \cite{ZhuH05, KSAKW06} or incorporate penalty functions. In this work, due to the high computational cost associated with iterative methods, we focus solely on the penalty-based approach for sparse coefficient selection.

\subsection{The $L_1$-norm RKLR model}
The $L_1$-norm penalty, widely known as Lasso regularization, is highly valued in feature selection for its computational efficiency and ability to produce sparse models \cite{Tibs96}. Its convexity also enables the use of scalable optimization techniques such as coordinate descent, making it suitable even for large-scale problems \cite{HaTW15}.

Adding $L_1$-norm penalty to the KLR model \eqref{eq.KLR3} can produce a sparse selection of the coefficient vector, and subsequently we get the following  $L_1$-norm RKLR model
\begin{equation}
	\label{eq.PKLR1}
	\min_{\boldsymbol{\alpha}\in \mathbb{R}^n} \frac{1}{n} \boldsymbol{1}^\top \ln\left(\boldsymbol{1}+\exp(-\boldsymbol{y}\odot\boldsymbol{K\alpha})\right)+ \frac{\lambda}{2}\boldsymbol{\alpha}^\top\boldsymbol{K\alpha} +\lambda_1\left\|\boldsymbol{\alpha}\right\|_1,
\end{equation}
where $\lambda_1$ is the sparsity regularization parameter. Although the $L_1$ norm penalty preserves the convexity of the KLR model, it results in an objective function that is no longer continuously differentiable.

Numerous other penalty functions can also induce sparsity in the coefficient vector, such as the $L_{1/2}$ norm \cite{XPJ13}, $L_0$ norm \cite{LDDS11}, and smoothly clipped absolute deviation (SCAD) \cite{FanL01}. However, these penalties are typically nonconvex, which means that the problem \eqref{eq.PKLR1} no longer guarantees a unique solution.

\subsection{The IKLR model}
To introduce the IKLR model proposed by \cite{LHGYS18}, we first recall the definition of Kre\u{\i}n space.
\begin{definition}[\cite{Bogn12}]
\label{def.KreinS}
An inner product space is a Kre\u{\i}n space $\mathcal{H_\mathcal{K}}$ if there exist two Hilbert spaces $\mathcal{H}_+$ and $\mathcal{H}_-$ such that: (1) all $f\in\mathcal{H_\mathcal{K}}$ can be decomposed into $f=f_+ + f_-$, where $f_+\in\mathcal{H}_+ $ and $f_-\in\mathcal{H}_-$, respectively, and (2) $\forall {f}$, $g\in\mathcal{H_\mathcal{K}}$, $\langle f, g\rangle_\mathcal{H_\mathcal{K}}=\langle f_+, g_+ \rangle_\mathcal{H_+} - \langle f_-, g_- \rangle_\mathcal{H_-}$.
\end{definition}
The function space generated by an indefinite kernel is a RKKS  rather than RKHS \cite{Bogn12}. If $\mathcal{H}_+$ and $\mathcal{H}_-$ are RKHSs, then $\mathcal{H_\mathcal{K}}$ is an RKKS with a unique indefinite kernel $\mathcal{K}$. Unlike the conventional RKHS for PD kernels, the inner products in the RKKS can be negative. Thus, the IKLR model with an indefinite kernel $\mathcal{K}$ is given by
\begin{equation}
	\label{eq.IKLR1}
	\min_{f\in\mathcal {H_\mathcal K}} \frac{1}{n} \sum_{i=1}^{n} \ln(1+e^{-y_if(\boldsymbol{x}_i)})+\frac{\lambda}{2}\left\|f\right\|_\mathcal{H_\mathcal{K}}^2,
\end{equation}
where $\mathcal {H_\mathcal K}$ is the RKKS generated by the indefinite kernel $\mathcal{K}(\cdot,\cdot)$. By the Representer Theorem in RKKS \cite{OMCS04}, the optimal discriminant function $f$ can be written as
\begin{eqnarray*}
	f=\sum_{i=1}^{n} \alpha_i\mathcal K(x_i,\cdot).
\end{eqnarray*}
Substituting this form into \eqref{eq.IKLR1} yields
\begin{equation}
	\label{eq.IKLR2}
	\min_{\boldsymbol{\alpha}\in\mathbb{R}^n} \frac{1}{n} \boldsymbol{1}^\top \ln(\boldsymbol{1}+\exp(-\boldsymbol{y}\odot\boldsymbol{K\alpha})) +\frac{\lambda}{2}\boldsymbol{\alpha}^\top\boldsymbol{K\alpha},
\end{equation}
where $\boldsymbol{K}$ is an indefinite kernel matrix. The indefiniteness of $\boldsymbol{K}$ makes the learning problem nonconvex. To solve this type of nonconvex optimization problem, we rely on the following result.
\begin{lemma}[\cite{OMCS04}]
	\label{Lem.Eigen}
	In the RKKS, there exists a positive decomposition of the indefinite kernel $\mathcal K{\boldsymbol{(x_i, x_j)}}$ such that
	\begin{equation*}
		\mathcal K{(\boldsymbol{x}_i, \boldsymbol{x}_j)}=\mathcal K_+{(\boldsymbol{x}_i, \boldsymbol{x}_j)}-\mathcal K_-{(\boldsymbol{x}_i, \boldsymbol{x}_j)}, \quad \forall \boldsymbol{x}_i, \boldsymbol{x}_j\in \mathcal{X},
	\end{equation*}
	where $\mathcal K_+$ and $\mathcal K_-$ are two PD kernels.
\end{lemma}

Using Lemma~\ref{Lem.Eigen} and following \cite{LHGYS18}, \eqref{eq.IKLR2} can be reformulated as
\begin{equation}
	\label{eq.IKLR3}
	\min_{\boldsymbol{\alpha}\in\mathbb{R}^n} \frac{1}{n} \boldsymbol{1}^\top \ln(\boldsymbol{1}+\exp(-\boldsymbol{y}\odot\boldsymbol{K}\boldsymbol{\alpha})) +\frac{\lambda}{2}\boldsymbol{\alpha}^\top\boldsymbol{(K_+-K_-)}\boldsymbol{\alpha},
\end{equation}
where $\boldsymbol{K_+}$ and $\boldsymbol{K_-}$ are two PSD kernel matrices obtained by the positive decomposition of $\boldsymbol{K}$.
Specifically, let the eigenvalue decomposition of the kernel matrix $\boldsymbol{K}$ be given by $\boldsymbol{K}=\boldsymbol{U \Lambda U^\top},$
where $\boldsymbol{\Lambda}=\mathrm{diag}(\mu_1,\mu_2,\cdots,\mu_n)$ is the diagonal matrix of eigenvalues satisfying $\mu_1 \geq \mu_2 \geq\cdots \geq \mu_n$, and $\boldsymbol{U}$ is an orthogonal matrix. Without loss of generality, we assume that the first $m$ eigenvalues are non-negative and the remaining $n-m$ eigenvalues are negative. Then, the positive decomposition of $\boldsymbol{K}$ is given by
\begin{equation}
\label{eq.IKLR6}
   \boldsymbol{K} = \boldsymbol{K_+} + \boldsymbol{K_-},
\end{equation}
where
\begin{eqnarray*}
\left\{
  \begin{array}{ll}
    \boldsymbol{K_+}=\boldsymbol{U} \mathrm{diag}(\mu_1+\tau,\cdots,\mu_m +\tau,\tau,\cdots,\tau) \boldsymbol{U^\top}, & \hbox{} \\
    \boldsymbol{K_-}=\boldsymbol{U} \mathrm{diag}(\tau,\cdots,\tau,\tau-\mu_{m+1},\cdots,\tau-\mu_n) \boldsymbol{U^\top}, & \hbox{}
  \end{array}
\right.
\end{eqnarray*}
and $\tau$ is a positive real number guaranteeing both $\boldsymbol{K_+}$ and $\boldsymbol{K_-}$ are positive definite. With this positive decomposition, the objective function \eqref{eq.IKLR3} can be expressed as a difference of two convex functions
\begin{eqnarray*}
	\min_{\boldsymbol{\alpha}\in\mathbb{R}^n} g(\boldsymbol{\alpha}) - h(\boldsymbol{\alpha}),
\end{eqnarray*}
where
\begin{eqnarray*}
\left\{
  \begin{array}{ll}
    g(\boldsymbol{\alpha})=\frac{1}{n} \boldsymbol{1}^\top \ln(\boldsymbol{1}+\exp(-\boldsymbol{y}\odot\boldsymbol{K}\boldsymbol{\alpha})) +\frac{\lambda}{2}\boldsymbol{\alpha}^\top\boldsymbol{K_+}\boldsymbol{\alpha}, & \hbox{} \\
 h(\boldsymbol{\alpha})=\frac{\lambda}{2}\boldsymbol{\alpha}^\top \boldsymbol{K_-}\boldsymbol{\alpha}.
 & \hbox{}
  \end{array}
\right.
\end{eqnarray*}
\begin{remark}
\rm
We briefly summarize the differences among the three models discussed above. KLR is a classification model that extends classical logistic regression by incorporating kernel functions. However, standard KLR does not promote sparsity in the solution. The $L_1$-norm RKLR extends KLR by introducing an $L_1$-norm penalty to encourage sparsity and feature selection. The IKLR model generalizes KLR by allowing the use of indefinite kernels, thereby providing greater flexibility in practical applications. However, IKLR does not inherently promote sparsity in the solution, which is the main focus of this paper.
\end{remark}

\section{The RIKLR model}
\label{sec:PIKLR}
This section introduces the proposed sparse indefinite kernel logistic regression model with $L_1$-norm regularization (RIKLR). We first present the model formulation, followed by an efficient optimization algorithm based on the proximal linearized approach and its convergence analysis.

\subsection{The $L_1$-norm RIKLR model}
\label{sec:Model}

As discussed in the aforementioned sections, the $L_1$-norm penalty is a powerful tool to produce sparse representation of the solution and has many efficient numerical algorithm supports \cite{CanRT06, Tibs96}.
To obtain a sparse coefficient vector $\boldsymbol{\alpha}$ for the indefinite kernel, we integrate $L_1$-norm penalty into \eqref{eq.IKLR2} and get the proposed $L_1$-norm RIKLR model as follows
\begin{equation}
	\label{eq.PIKLR1}
	\min_{\boldsymbol{\alpha}\in\mathbb{R}^n} \frac{1}{n} \boldsymbol{1}^\top \ln(\boldsymbol{1}+\exp(-\boldsymbol{y}\odot\boldsymbol{K\alpha}))+ \frac{\lambda}{2}\boldsymbol{\alpha}^\top\boldsymbol{K\alpha}+\lambda_1\left\|\boldsymbol{\alpha}\right\|_1,
\end{equation}
where $\boldsymbol{K}$  is an indefinite kernel matrix, and $\lambda$, $\lambda_1>0$ are regularization parameters controlling smoothness and sparsity, respectively. The $L_1$-norm regularization introduces nonsmoothness, rendering the optimization problem nonconvex and nondifferentiable.

To solve the $L_1$-norm RIKLR problem \eqref{eq.PIKLR1}, with the same strategy for solving \eqref{eq.IKLR2} we can transform \eqref{eq.PIKLR1} into
\begin{equation}
	\label{eq.PIKLR2}
	\min_{\boldsymbol{\alpha}\in\mathbb{R}^n} \frac{1}{n} \boldsymbol{1}^\top \ln(\boldsymbol{1}+\exp(-\boldsymbol{y}\odot\boldsymbol{K}\boldsymbol{\alpha}))+ \frac{\lambda}{2}\boldsymbol{\alpha}^\top\boldsymbol{(K_+-K_-)}\boldsymbol{\alpha}+ \lambda_1\left\|\boldsymbol{\alpha}\right\|_1,
\end{equation}
where $\boldsymbol{K_+}$ and $\boldsymbol{K_-}$ are two PSD kernel matrices.  Then, the objective function of \eqref{eq.PIKLR2} can be written as the following DC functions
\begin{equation}
	\label{eq.IKLR5}
	\min_{\boldsymbol{\alpha}\in\mathbb{R}^n} f(\boldsymbol{\alpha}):=  g(\boldsymbol{\alpha}) - h(\boldsymbol{\alpha}),
\end{equation}
where
\begin{equation}
\label{eq.PIKLR3}
\left\{
  \begin{array}{ll}
  g(\boldsymbol{\alpha})=\frac{1}{n} \boldsymbol{1}^\top \ln(\boldsymbol{1}+\exp(-\boldsymbol{y}\odot\boldsymbol{K}\boldsymbol{\alpha})) +\frac{\lambda}{2}\boldsymbol{\alpha}^\top\boldsymbol{K_+}\boldsymbol{\alpha} +\lambda_1\left\|\boldsymbol{\alpha}\right\|_1, & \hbox{} \\
  h(\boldsymbol{\alpha})=\frac{\lambda}{2}\boldsymbol{\alpha}^\top\boldsymbol{K_-}\boldsymbol{\alpha},
 & \hbox{}
  \end{array}
\right.
\end{equation}
and $g(\boldsymbol{\alpha})$ is nonsmooth.
Furthermore, for the PSD kernel matrix $\boldsymbol{K_+}$, we can get its eigenvalue decomposition
$$\boldsymbol{K_+}=\boldsymbol{U\Lambda_{1} U^\top},$$
where $\boldsymbol{\Lambda_{1}}=\mathrm{diag}(\mu_1+\tau,\cdots,\mu_m +\tau,\tau,\cdots,\tau)$ is the diagonal matrix and $\boldsymbol{U}$ is an orthogonal matrix. If we set
$$\boldsymbol{B}=\boldsymbol{\sqrt{\Lambda_{1}} U^\top}$$
with $\boldsymbol{\sqrt{\Lambda_{1}}} = \mathrm{diag}(\sqrt{\mu_1+\tau},\cdots,\sqrt{\mu_m +\tau},\sqrt{\tau}, \cdots ,\sqrt{\tau})$, then we have
$$\boldsymbol{K_+}=\boldsymbol{U\sqrt{\Lambda_{1}} \sqrt{\Lambda_{1}} U^\top}=\boldsymbol{B^\top B}.$$
Thus, \eqref{eq.PIKLR3} can be further written as
\begin{equation}
	\label{eq.SIKLR4}
\left\{
  \begin{array}{ll}
  g(\boldsymbol{\alpha})=\frac{1}{n} \boldsymbol{1}^\top \ln\left(\boldsymbol{1}+\exp(-\boldsymbol{y}\odot\boldsymbol{K}\boldsymbol{\alpha})\right)+ \frac{\lambda}{2}\left\|\boldsymbol{B\alpha}\right\|^2+ \lambda_1\left\|\boldsymbol{\alpha}\right\|_1,& \hbox{} \\
  h(\boldsymbol{\alpha})=\frac{\lambda}{2}\boldsymbol{\alpha}^\top\boldsymbol{K_-}\boldsymbol{\alpha}.
 & \hbox{}
  \end{array}
\right.
\end{equation}

\begin{remark}\rm
The main difference between the IKLR model~\eqref{eq.IKLR2} and the proposed $L_1$-norm RIKLR model~\eqref{eq.PIKLR1} lies in the introduction of an $L_1$-norm penalty to promote sparsity in the solution.
Unlike the smooth regularization used in IKLR, the $L_1$-norm penalty introduces nonsmoothness, rendering the optimization algorithms such as CCCP and CCICP in \cite{LHGYS18} inapplicable.
To address this challenge, we develop a proximal linearized algorithm (PLA) that directly handles the nonsmooth convex term $g(\boldsymbol{\alpha})$ induced by the $L_1$-norm.
This extension not only enables the use of more general sparsity-inducing regularizers but also broadens the applicability of the method to a wider class of nonconvex kernel learning problems.
\end{remark}

\subsection{Optimization algorithm}\label{sec:algorithm}

The objective function of the $L_1$-norm RIKLR model \eqref{eq.PIKLR1} is expressible as a DC functions as in \eqref{eq.IKLR5}, thereby falling within the DC optimization framework \cite{LeDin18}. DC optimization algorithms have proven effective for addressing a variety of structured and practical problems \cite{AnTao05,HoTu99,ChenH98}. Notably, Sun et al. \cite{SunSC03} developed a proximal point algorithm for minimizing DC functions by leveraging the convex properties of their components, while Souza et al. \cite{SouOS16} established global convergence guarantees for a proximal linearized algorithm applied to nonsmooth DC minimization. Inspired by these advances, we employ the proximal linearized algorithm (PLA) proposed by \cite{SouOS16} to solve our nonsmooth DC optimization problem.

According to \cite{SouOS16}, there are two main steps to produce the iterative sequence. Therefore, in this part we just present how to constructing the sequence $\{\boldsymbol{\alpha}_k\}$ with these two steps. In the first step, given the current $k$-th iteration of the solution $\boldsymbol{\alpha}_k$, we compute the subgredient
$$\boldsymbol{\omega}_{k} \in \partial h(\boldsymbol{\alpha}_k)=\left\{\lambda\boldsymbol{K_-}\boldsymbol{\alpha}_k\right\},$$
where $\partial h(\boldsymbol{\alpha}_k)$ denotes the subdifferential of $h$ at $\boldsymbol{\alpha}_k$. Because $h$ in \eqref{eq.IKLR5} is differentiable, we directly get $\boldsymbol{\omega}_{k} = \lambda\boldsymbol{K_-}\boldsymbol{\alpha}_k$. In the second step,  with linearization technique,  we can construct the following function
\begin{equation}
	\label{eq.algorithm2}
	\tilde{f}(\boldsymbol{\alpha}) = g(\boldsymbol{\alpha})-\boldsymbol{\omega}_{k}^\top (\boldsymbol{\alpha}-\boldsymbol{\alpha}_k) +\frac{1}{2\gamma_k}\left\|\boldsymbol{\alpha}-\boldsymbol{\alpha}_k\right\|^2,
\end{equation}
where $\gamma_k$ is a bounded sequence of positive numbers such that $\lim \inf_k \gamma_k > 0$. Then, $\boldsymbol{\alpha}_{k+1}$ is obtained by solving the nonsmooth optimization problem
\begin{eqnarray}
	\label{eq.algorithm4}
	\boldsymbol{\alpha}_{k+1} \in \mathop\mathrm{argmin}_{\boldsymbol{\alpha}\in \mathbb{R}^n} \tilde{f}(\boldsymbol{\alpha}) &=& \frac{1}{n} \boldsymbol{1}^\top \ln(\boldsymbol{1}+\exp(-\boldsymbol{y}\odot\boldsymbol{K}\boldsymbol{\alpha})) +\frac{\lambda}{2}\left\|\boldsymbol{B\alpha}\right\|_2^2 +\lambda_1\left\|\boldsymbol{\alpha}\right\|_1\nonumber\\
 & &-\boldsymbol{\omega}_{k}^\top (\boldsymbol{\alpha}-\boldsymbol{\alpha}_k) +\frac{1}{2\gamma_k}\left\|\boldsymbol{\alpha}-\boldsymbol{\alpha}_k\right\|^2.
\end{eqnarray}
Iterate the above two main steps until the solution sequence $\{\boldsymbol{\alpha}_k\}$ satisfies a certain convergence criterion, and the stop point $\boldsymbol{\alpha^*}$ is taken as the final solution. Summarizing the above discussion, we present the details of the proposed algorithm for solving \eqref{eq.IKLR5} in the Algorithm~\ref{alg1}.

After obtaining the optimal $\boldsymbol{\alpha^*}$ through Algorithm~\ref{alg1}, we can predict the label of a test data point $\boldsymbol{z}$ by computing
\begin{equation*}	p(\boldsymbol{z})=\frac{\exp(\boldsymbol{K_{z}\alpha^*})}{1+\exp(\boldsymbol{K_{z}\alpha^*})}
\end{equation*}
with $\boldsymbol{K_{z}} = \left[\mathcal K{(\boldsymbol{x}_1, \boldsymbol{z})}, \mathcal K{(\boldsymbol{x}_2, \boldsymbol{z})},\cdots, \mathcal K{(\boldsymbol{x}_n, \boldsymbol{z})}\right]$.
If $p(z) \geq 0.5$, then the predicted label $y^*$ is assigned as $+1$, otherwise it is assigned as $0$. Furthermore, note that the main computational challenge of Algorithm~\ref{alg:algorithm1} lies in solving the nonsmooth optimization problem \eqref{eq.algorithm4}. With some algebraic manipulation, it can be shown that $\tilde{f}(\boldsymbol{\alpha})$ is a convex and nonsmooth optimization problem and can be easily solved with the \texttt{CVXR} R package \cite{FuNB20}, which provides an object-oriented modeling language for convex optimization and has found wide and easy applications for users from various disciplines.

\begin{algorithm}[]\label{alg1}
	\caption{A PLA for the $L_1$-norm RIKLR model \eqref{eq.IKLR5}.}
	\label{alg:algorithm1}
	\KwIn{Data matrix $\mathbb{X}_{p\times n}$, label vector $\boldsymbol{y}$, kernel function $\mathcal{K}(\cdot,\cdot)$, tuning parameters $\lambda$ and $\lambda_1$, $\tau$, initial coefficient vector $\alpha_0$ and sequence $\{\gamma_k\}$.}
	\KwOut{the coefficient vector $\boldsymbol{\alpha^*}$.}
	\BlankLine
	
	
	Compute indefinite kernel matrix $\boldsymbol{K}$ and its positive decomposition $\boldsymbol{K_+}$ and $\boldsymbol{K_-}$;

    Construct DC decomposition: $f(\boldsymbol{\alpha})=g(\boldsymbol{\alpha})-h(\boldsymbol{\alpha})$;
	
  \Repeat{Certain stop criterion satisfied}{Compute $\boldsymbol{\omega}_{k} = \lambda\boldsymbol{K_-}\boldsymbol{\alpha}_k$;\\
         Update $\boldsymbol{\alpha}$ by solving  \eqref{eq.algorithm4} to obtain  $\boldsymbol{\alpha}_{k+1}$;}
	Output the sparse solution $\boldsymbol{\alpha^*}=\boldsymbol{\alpha}_{k+1}$.
\end{algorithm}

\subsection{Convergence analysis}\label{sec:convergence}

This section establishes the convergence properties of Algorithm~\ref{alg1}.
The analysis follows the general framework in~\cite{SouOS16},
which studies a broad class of DC optimization problems.
We adapt their results to the proposed nonsmooth and nonconvex objective function.

\begin{theorem}
\label{Thm:Decent}
Let $\{\boldsymbol{\alpha}_k\}$ be the sequence generated by Algorithm~\ref{alg1} and suppose  $\{\boldsymbol{\alpha}_k\}$ is bounded. Assume the step-size sequence $\{\gamma_k\}$ satisfies $\gamma_k > 0$ and $\liminf_{k \to \infty} \gamma_k > 0$. Then the following holds:
\begin{enumerate}
	\item The algorithm either stops at a critical point of $f$;
	\item Or, every cluster point of $\{\boldsymbol{\alpha}_k\}$ is the critical point of the DC function $f$.
\end{enumerate}
\end{theorem}
\begin{proof}
Recall that $f(\boldsymbol{\alpha}) = g(\boldsymbol{\alpha}) - h(\boldsymbol{\alpha})$ is a DC function, where $g$ is convex and $h$ is convex and differentiable. From Algorithm~\ref{alg1}, we have $\boldsymbol{\omega}_k = \nabla h(\boldsymbol{\alpha}_k) = \lambda \boldsymbol{K}_- \boldsymbol{\alpha}_k$. The first order necessary condition of nonsmooth convex problem \eqref{eq.algorithm2} gives
\begin{equation}
\label{eq:fnecessary}
\boldsymbol{\omega}_{k}\in \partial g(\boldsymbol{\alpha}_{k+1})+ \frac{1}{\gamma_k} (\boldsymbol{\alpha}_{k+1}-\boldsymbol{\alpha}_k).
\end{equation}
If $\boldsymbol{\alpha}_{k+1} = \boldsymbol{\alpha}_k$, then \eqref{eq:fnecessary} implies $\boldsymbol{\omega}_k \in \partial g(\boldsymbol{\alpha}_k)$. Since $\boldsymbol{\omega}_k = \nabla h(\boldsymbol{\alpha}_k)$, it follows that $\nabla h(\boldsymbol{\alpha}_k) \in \partial g(\boldsymbol{\alpha}_k)$, so $\boldsymbol{\alpha}_k$ is a critical point of $f$.

Now suppose $\boldsymbol{\alpha}_{k+1} \neq \boldsymbol{\alpha}_k$. By the convexity and differentiability of $h$, we have
\begin{equation}
h(\boldsymbol{\alpha}_{k+1})\geq h(\boldsymbol{\alpha}_{k}) + \boldsymbol{\omega}_{k}^\top (\boldsymbol{\alpha}_{k+1}-\boldsymbol{\alpha}_k). \label{con:p1}
\end{equation}
From the optimality of $\boldsymbol{\alpha}_{k+1}$ in \eqref{eq.algorithm4}, the following inequality holds
\begin{equation}
g(\boldsymbol{\alpha}_{k})\geq g(\boldsymbol{\alpha}_{k+1}) - \boldsymbol{\omega}_{k}^\top (\boldsymbol{\alpha}_{k+1}-\boldsymbol{\alpha}_k) +\frac{1}{2\gamma_k}\left\|\boldsymbol{\alpha}_{k+1}-\boldsymbol{\alpha}_k\right\|^2. \label{con:p2}
\end{equation}
Taking the sum of the inequalities \eqref{con:p1} and \eqref{con:p2} leads to
\begin{equation}
\label{con:p3}
f(\boldsymbol{\alpha}_{k})\geq f(\boldsymbol{\alpha}_{k+1}) +\frac{1}{2\gamma_k}\left\|\boldsymbol{\alpha}_{k+1}-\boldsymbol{\alpha}_k\right\|^2.
\end{equation}
Since $\{\gamma_k\}$ is a positive sequence and $\boldsymbol{\alpha}_{k+1} \neq \boldsymbol{\alpha}_k$, we have $f(\boldsymbol{\alpha}{k+1}) < f(\boldsymbol{\alpha}_k)$, so the algorithm is strictly decreasing.
From \eqref{eq.PIKLR1}, the inequality $f(\boldsymbol{\alpha})\geq \frac{\lambda}{2}\boldsymbol{\alpha}^\top\boldsymbol{K\alpha}$ holds. Note that the kernel matrix $\boldsymbol{K}$ is finite-dimensional and has bounded entries, its minimum eigenvalue is finite, and thus $f$ is bounded below.
Since $\{\boldsymbol{\alpha}_{k}\}$ is bounded and $f$ is continuous, we can get $\lim_{k\rightarrow\infty}f(\boldsymbol{\alpha}_{k}) = f(\boldsymbol{\alpha}^*)$ for some cluster point $\boldsymbol{\alpha}^*$ of $\{\boldsymbol{\alpha}_{k}\}$.

To prove every cluster point $\boldsymbol{\alpha}^*$ is the critical point of $f$, we first set $\boldsymbol{\omega}^*$ be the cluster point of $\{\boldsymbol{\omega}_k\}$. Then, there exist two subsequences $\{\boldsymbol{\alpha}_{k_j}\}$ and $\{\boldsymbol{\omega}_{k_j}\}$ converging to $\boldsymbol{\alpha}^*$ and $\boldsymbol{\omega}^*$, respectively. The definition of $\boldsymbol{\omega}_k$ directly shows that $\boldsymbol{\omega}^* = \nabla h(\boldsymbol{\alpha^*})$. According to \eqref{eq:fnecessary}, there exists $\boldsymbol{z}_{k_{j+1}}\in \partial g(\boldsymbol{\alpha}_{k_{j+1}})$ such that
\begin{eqnarray*}
  \|\boldsymbol{\omega}_{k_j} - \boldsymbol{z}_{k_{j+1}}\| = \frac{1}{\lambda_{k_j}}\|\boldsymbol{\alpha}_{k_{j+1}} - \boldsymbol{\alpha}_{k_j}\|.
\end{eqnarray*}
With \eqref{con:p3}, for the subsequence $\{\boldsymbol{\alpha}_{k_j}\}$ we have
$\sum_{j=0}^{n-1} \frac{1}{2\gamma_{k_j}}\left\|\boldsymbol{\alpha}_{k_{j+1}}-\boldsymbol{\alpha}_{k_j}\right\|^2 \leq f(\boldsymbol{\alpha}_{k_0}) - f(\boldsymbol{\alpha}_{k_n}). \label{con:p4}
$
Since $\{\gamma_k \}$ is bounded, it follows that
$\sum_{j=0}^{\infty} \left\|\boldsymbol{\alpha}_{k_{j+1}}-\boldsymbol{\alpha}_{k_j}\right\|^2 < \infty,
$
which implies
\begin{eqnarray*}
  \lim \limits_{j\rightarrow \infty} \left\|\boldsymbol{\alpha}_{k_{j+1}} - \boldsymbol{\alpha}_{k_j}\right\|=0,\text{ and } \lim\limits_{j\rightarrow \infty}\boldsymbol{\omega}_{k_j} = \lim\limits_{j\rightarrow \infty}\boldsymbol{z}_{k_{j+1}} = \boldsymbol{\omega}^*.
\end{eqnarray*}
By the closedness of the subdifferential mapping of $g$ and  $\boldsymbol{z}_{k_{j+1}}\in \partial g(\boldsymbol{\alpha}_{k_{j+1}})$, we have $\boldsymbol{\omega}^* = \nabla h(\boldsymbol{\alpha^*}) \in \partial g(\boldsymbol{\alpha}^*)$, and thus $\boldsymbol{\alpha}^*$ is a critical point of $f$.
\end{proof}

Although the proposed algorithm strictly follows the optimization framework established in \cite{SouOS16}, its \emph{global convergence} cannot be derived in the same manner.  Nevertheless, it follows from \cite{AtBS13} that $f(\boldsymbol{\alpha})$ is a \emph{Kurdyka--{\L}ojasiewicz (KL)} function and satisfies the KL property at the limit point $\boldsymbol{\alpha}^\ast$ with exponent $\theta = 1/2$. By invoking the standard KL-based convergence analysis in \cite{AtBS13}, the \emph{local linear convergence rate} of the proposed Algorithm~\ref{alg:algorithm1} can be rigorously established. Here, we only outline the key steps of the proof and refer interested readers to \cite{AtBS13} for detailed analysis.

\begin{theorem}
\label{thm:local_linear_liminf}
Let the setup and notation be the same as that in Theorem~\ref{Thm:Decent}. For the step-size sequence $\{\gamma_k\}$, we set $\gamma_{\min}:=\inf_{k\ge0}\gamma_k>0$ and $\gamma_{\max}:=\sup_{k\ge0}\gamma_k<\infty$.
Then there exist constants $C>0$ and $m\in(0,1)$ such that
for all sufficiently large $k$,
\begin{equation}
\label{eqThm2}
\|\boldsymbol{\alpha}_k-\boldsymbol{\alpha}^\ast\|\le C\,m^{\,k},
\end{equation}
where $m = \Big(1+\frac{2}{\gamma_{\max}\,c_\varphi^{2}
    \big(L_h+\tfrac{1}{\gamma_{\min}}\big)^{2}}\Big)^{-1/2}$,  $L_h = \lambda(\tau - \mu_n)$, and  $c_\varphi>0$ is the desingularizing constant in the KL desingularizer
$\varphi(s)=c_\varphi s^{1-\theta}$.
\end{theorem}

\begin{proof}
The proof follows the standard KL-based outline. Firstly, by Theorem~\ref{Thm:Decent} and \eqref{con:p3}
we directly show that a sufficient decrease appears in iteration
\begin{eqnarray}
\label{eqthm321}
f(\boldsymbol{\alpha}_k)-f(\boldsymbol{\alpha}_{k+1})
\ge \frac{1}{2\gamma_k}\|\boldsymbol{\alpha}_{k+1}-\boldsymbol{\alpha}_k\|^2
\ge \frac{1}{2\gamma_{\max}}\|\boldsymbol{\alpha}_{k+1}-\boldsymbol{\alpha}_k\|^2.
\end{eqnarray}
Secondly, by the optimality condition \eqref{eq:fnecessary}, there exists a $\boldsymbol{z}_{k+1}\in \partial g(\boldsymbol{\alpha}_{k+1})$ such that
\begin{equation*}
  \nabla h(\boldsymbol{\alpha}_{k}) = \boldsymbol{z}_{k+1} + \tfrac{1}{\gamma_k}(\boldsymbol{\alpha}_{k+1}-\boldsymbol{\alpha}_k).
\end{equation*}
Because $\partial f(\boldsymbol{\alpha}_{k+1})
   = \partial g(\boldsymbol{\alpha}_{k+1}) - \nabla h(\boldsymbol{\alpha}_{k+1})$. We set $\mathbf{s}_{k+1}=\mathbf{z}_{k+1}-\nabla h(\boldsymbol{\alpha}_{k+1})
   \in \partial f(\boldsymbol{\alpha}_{k+1})$. Thus, we obtain
\[
  \mathbf{s}_{k+1}
  = \nabla h(\boldsymbol{\alpha}_k)-\nabla h(\boldsymbol{\alpha}_{k+1})
    - \tfrac{1}{\gamma_k}(\boldsymbol{\alpha}_{k+1}-\boldsymbol{\alpha}_k).
\]
Taking norms and applying the Lipschitz continuity of $\nabla h$ gives the following relative-error bound
\begin{equation}\label{eq:rel_error}
  \operatorname{dist}\big(\mathbf{0},\partial f(\boldsymbol{\alpha}_{k+1})\big)
  \le \|\mathbf{s}_{k+1}\|
  \le \bar{c}\|\boldsymbol{\alpha}_{k+1}-\boldsymbol{\alpha}_k\|,
\end{equation}
where $\bar{c} = L_h+\tfrac{1}{\gamma_{\min}}$,  $\operatorname{dist}\big(\mathbf{0},\partial f(\boldsymbol{\alpha}_{k+1})\big)= \inf_{\mathbf{v}\in\partial f(\boldsymbol{\alpha})}\|\mathbf{v}\|$ denotes the Euclidean distance from $\mathbf{0}$ to the set $\partial f(\boldsymbol{\alpha}_{k+1})$, and $L_h = \lambda\|\boldsymbol{K}_-\| = \lambda(\tau - \mu_n)$ is the Lipschitz constant of $\Delta h(\boldsymbol{\alpha})$.

%
%
%
%
%

By the KL property at $\boldsymbol{\alpha}^\ast$ (exponent parameter $\theta = 1/2$) there exists $c_\varphi>0$ and a neighborhood
$U$ such that for large $k$ with $\boldsymbol{\alpha}_{k+1}\in U$,
\[
c_\varphi(\tfrac12)\,(f(\boldsymbol{\alpha}_{k+1})-f(\boldsymbol{\alpha}^*))^{-1/2}
\operatorname{dist}\big(\mathbf{0},\partial f(\boldsymbol{\alpha}_{k+1})\big)\ge 1.
\]
Combining with the relative-error bound \eqref{eq:rel_error} yields (for $k\ge K$ sufficiently large)
\begin{equation}\label{eqthm325}
 \|\boldsymbol{\alpha}_{k+1}-\boldsymbol{\alpha}_k\|
\ge \frac{2}{\bar c\,c_\varphi}\,(f(\boldsymbol{\alpha}_{k+1})-f(\boldsymbol{\alpha}^*))^{1/2}.
\end{equation}
Substitute \eqref{eqthm325} into
the sufficient-decrease bound \eqref{eqthm321} to obtain, for all large $k$,
\begin{equation}
f(\boldsymbol{\alpha}_k)-f(\boldsymbol{\alpha}_{k+1})
\ge \frac{1}{2\gamma_{\max}}\Big(\frac{2}{\bar c\,c_\varphi}\Big)^2
\big(f(\boldsymbol{\alpha}_{k+1})-f(\boldsymbol{\alpha}^*)\big).
\end{equation}
Set
\[
B := \frac{1}{2\gamma_{\max}}\Big(\frac{2}{\bar c\,c_\varphi}\Big)^2
= \frac{2}{\gamma_{\max}\,c_\varphi^{2}\,\bar c^{2}}>0.
\]
Then $f(\boldsymbol{\alpha}_k)-f(\boldsymbol{\alpha}^*) \ge(1+B)(f(\boldsymbol{\alpha}_{k+1})-f(\boldsymbol{\alpha}^*))$,
so $f(\boldsymbol{\alpha}_{k+1})-f(\boldsymbol{\alpha}^*)\le q\,(f(\boldsymbol{\alpha}_k)-f(\boldsymbol{\alpha}^*)$ with
$q:=1/(1+B)\in(0,1)$ for all large $k$.

According to the standard KL analysis in \cite{AtBS13}, geometric decay of
function values yields geometric decay of steps and hence of the iterates:
there exist $C>0$ and $m:=q^{1/2}\in(0,1)$ such that for all sufficiently large $k$,
\[
\|\boldsymbol{\alpha}_k-\boldsymbol{\alpha}^\ast\|\le C\,m^{\,k}.
\]
The explicit expression of $m$ in \eqref{eqThm2} follows from substituting
$\bar c = L_h + 1/\gamma_{\min}$ and $B$ above, which completes the proof.
\end{proof}

\begin{remark}
\rm
The positive-definite decomposition~\eqref{eq.IKLR6} ensures that
$g(\boldsymbol{\alpha})$ is strongly convex with modulus $\rho=\tau$,
and that $\nabla h(\boldsymbol{\alpha})$ is Lipschitz continuous
with constant $L=\lambda(\tau-\mu_n)$.
However, the condition $\rho>2L$ required for global convergence \cite{SouOS16}
cannot be satisfied due to the indefinite nature of $\boldsymbol{K}$ (where $\mu_n<0$). On the other hand, the object function $f(\boldsymbol{\alpha})$ is composed of logistic loss, the quadratic regularization term, and the $L_1$-norm penalty, which are all semi-algebraic (subanalytic) functions, and allows the use of KL framework \cite{AtBS13} to establish the local linear convergence rate of the proposed algorithm.
\end{remark}

\section{Numerical experiment}\label{sec:experiments}

In this section, we present the experimental results of applying the $L_1$-norm RIKLR model to binary classification problems and compare its performance with other related methods.

\subsection{Experimental setup}
Here, we first specify the kernel selection problem in our experiment. Since the proposed $L_1$-norm RIKLR model is designed for indefinite kernel and we can also note that when the PD kernel is used the proposed model will reduce to the $L_1$-norm RKLR model, we will use the indefinite and PD kernels. For the indefinite kernel, we choose the truncated $\ell_1$ distance (TL1) indefinite kernel, which is defined as
\begin{eqnarray*}
\mathcal{K}(\boldsymbol{x}_i, \boldsymbol{x}_j) = \max\{\eta - \|\boldsymbol{x}_i - \boldsymbol{x}_j\|_1, 0\}.
\end{eqnarray*}
As mentioned in \cite{HuSWHM17}, the TL1 kernel function is more robust to the parameter $\eta$ and typically set to $\eta = 0.7d$, where $d$ is the feature dimension.
For the PD kernel, we choose the radial basis function (RBF) kernel, which is defined as
\begin{eqnarray*}
\mathcal{K}(\boldsymbol{x}_i, \boldsymbol{x}_j) = \exp\left(-\frac{\|\boldsymbol{x}_i - \boldsymbol{x}_j\|_2^2}{\sigma^2}\right),
\end{eqnarray*}
where $\sigma$ is the kernel width. We integrate the aforementioned two different kernel functions to the related KLR models given in Table~\ref{table:11} and compare its performance.
\begin{table}[htp]
\footnotesize
\centering
\caption{KLR model and its kernel function.}\label{table:11}
\begin{tabular}{lr}
	\hline
	Model & Kernel function  \\
	\hline
	KLR & RBF kernel\\
	\hline
	$L_1$-norm RKLR&RBF kernel \\
	\hline
	IKLR &TL1 kernel\\
	\hline
	$L_1$-norm RIKLR&TL1 kernel\\		
	\hline
\end{tabular}
\end{table}

In our experiment, we select the values of tuning parameter $\lambda$ in all models , $\lambda_1$ in $L_1$-norm RKLR and $L_1$-norm RIKLR models, and the RBF kernel width $\sigma$ by fivefold cross-validation on the training set from the set $\{0.0001, 0.001, 0.01, 0.1, 1, 5, 10\}$. The same strategy was also used in \cite{LHGYS18}.
Inspired by the convergence analysis of Algorithm~\ref{alg:algorithm1}, a combined stop criteria
\begin{eqnarray*}\max\big(\left\|\boldsymbol{\alpha}_{k+1}-\boldsymbol{\alpha}_k\right\|, |f(\boldsymbol{\alpha}_k)-f(\boldsymbol{\alpha}_{k+1})| \big)< \epsilon
\end{eqnarray*}
will be used and $\epsilon$ usually set to be $10^{-4}$.
Note that $\tau$ denotes the minimum eigenvalue of $\boldsymbol{K}+$ and $\boldsymbol{K}-$, which may adversely affect the numerical stability and the practical convergence speed of the algorithm \cite{GoVL96}. Owing to the difficulty of theoretical analysis, we empirically recommend setting $\tau = 10^{-6}$.
Since $\gamma_k$ is a bounded sequence of positive numbers, i.e. $\lim\inf_k \ \gamma_k > 0$, we set $\gamma_k =1$ for simplicity.
To evaluate the performance of the models, we randomly split each dataset into two halves, with one half used for training and the other for testing. The process is repeated 10 times to generate 10 independent results for each dataset, and then the average results and its standard deviation are reported.

\subsection{Results on Kaggle and UCI Databases}

Here, we consider six different datasets from kaggle machine learning database and the UCI machine learning repository . We summarize the statistics of the datasets in Table~\ref{table:1} including the feature dimension $d$, the number of data points $n$, and the minimum and maximum eigenvalues $\mu_{\min}$ and $\mu_{\max}$ of the TL1 kernel.
It can be clearly seen that compared with other datasets, the negative eigenvalues of the kernel matrix of SPECT dataset are relatively large, which means the negative eigenvalues may contain some important information and cannot be ignored.
\begin{table}[htp]
\footnotesize
\centering
\caption{Statistics for six datasets.}\label{table:1}
\begin{tabular}{lrrrr}
	\hline
	Dataset & $d$ & $n$ &$\mu_{\min}$ &$\mu_{\max}$    \\
	\hline
	haberman &3 &306&-0.204&215.23\\
	\hline
	heart &8 &316 &0.006 &682.82 \\
	\hline
	fertility &9 &100&-0.249&162.17 \\
	\hline
	transfusion &4 &748&-0.319&836.35 \\
	\hline
	Wholesale &7 &440&0.002&903.99 \\
	\hline
	SPECT &22 &268&-12.01&932.39 \\		
	\hline
	\end{tabular}
\end{table}

Table~\ref{table:2} summarizes the experimental results on six datasets, and the best results are highlighted in bold. As the main purpose of introducing sparse solution, the number of selected features are also reported. The kernel coefficient $\alpha_i$ will be set to zero when it falls below the small threshold, which is $10^{-10}$.
\begin{table}[htp]
\footnotesize
\centering
\caption{Classification accuracy and the number of selected features.}\label{table:2}
\begin{tabular}{lrrrr}
	\hline
	Dataset & KLR & $L_1$-norm RKLR & IKLR & $L_1$-norm RIKLR \\
	\hline
	haberman & 0.733 $\pm$ 0.030 (153) & 0.729 $\pm$ 0.030 (45)&  0.727 $\pm$ 0.035 (153)& $\boldsymbol {0.736 \pm 0.019 (4)}$\\
	\hline
	heart   & 0.634 $\pm$ 0.022 (158) & 0.656 $\pm$ 0.035 (42)& 0.642 $\pm$ 0.031 (158)& $\boldsymbol { 0.682 \pm 0.034 (25)}$ \\
	\hline
	fertility   &  0.752 $\pm$ 0.023 (50) & 0.760 $\pm$ 0.021 (7)& 0.790 $\pm$ 0.064 (50) & $\boldsymbol { 0.882 \pm 0.022 (8)}$ \\
	\hline
	transfusion   &  0.741 $\pm$ 0.048 (374) & 0.748 $\pm$ 0.032 (290)& 0.769 $\pm$ 0.023 (374)& $\boldsymbol { 0.800 \pm 0.176 (196)}$  \\
	\hline
	Wholesale   &  0.681 $\pm$ 0.021 (220) & $\boldsymbol {0.685 \pm 0.030 (135) }$& 0.667  $\pm$ 0.017 (220)& 0.682 $\pm$ 0.023 (3) \\
	\hline
	SPECT   & 0.782 $\pm$ 0.041 (134) & 0.798 $\pm$ 0.019 (38)  & 0.792 $\pm$ 0.024 (134) & $\boldsymbol {0.838 \pm 0.021 (8)}$ \\
	\hline		
\end{tabular}
\end{table}
The proposed $L_1$-norm RIKLR consistently outperforms the other methods on most datasets.
Among the four methods, KLR is simple and fast, but it has the lowest classification accuracies on four of the six datasets.
In terms of classification accuracy, IKLR outperforms KLR on most datasets, but both of them tend to select too many features.
$L_1$-norm RKLR selects a smaller number of features, but it performs worse than IKLR on fertility and transfusion datasets.
Overall, the proposed $L_1$-norm RIKLR achieves higher classification accuracy than the other methods on most datasets with a much smaller number of features, indicating its effectiveness in obtaining sparse solutions.
Additionally, we find that the TL1 kernel used for training is still positive definite in several datasets, such as heart and Wholesale. In heart datasets, there is no significant difference in the classification accuracy between KLR and IKLR, and in Wholesale dataset the classification accuracy of the PD kernel method is better than the indefinite kernel method.

\subsection{Results on Mnist Database}

\begin{figure}[htp]
	\centering
	\includegraphics[width=0.8\textwidth,height=0.4\textwidth]{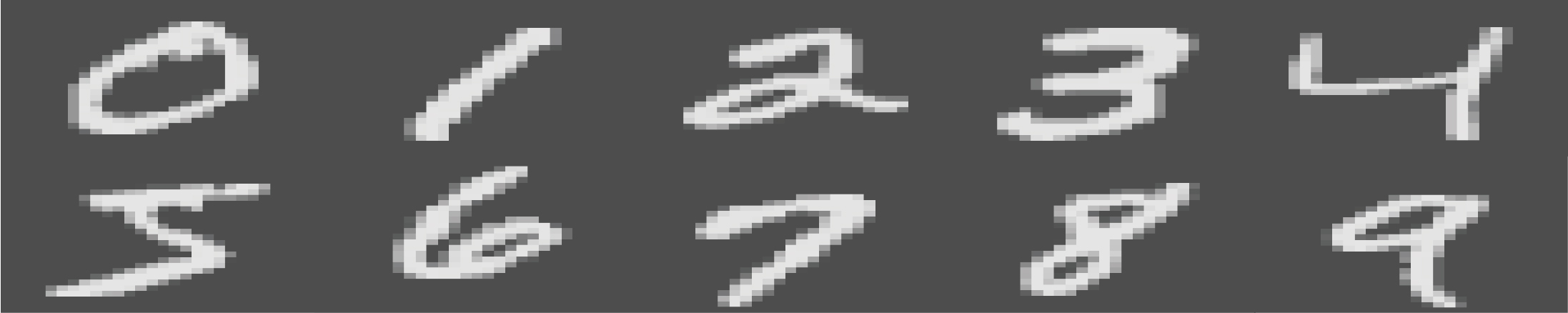}\\
	\caption{Some examples from the Mnist Database.}\label{Fig1}
\end{figure}

Apart from using the proposed $L_1$-norm RIKLR model on the kaggle and UCI database, we also illustrate the use of the TL1 kernel for handwritten digit recognition in the model.
Mnist database contains the 70,000 examples and labels (numbers from 0 to 9). Figure~\ref{Fig1} shows some image examples of the database.
During the experiment, due to the abundance of examples and to save time, we randomly selected only a portion of the examples for experiment.
Table \ref{table:3} presents the statistics including the minimum and maximum eigenvalues of the TL1 kernel, i.e., $\mu_{\min}$ and $\mu_{\max}$, the average classification accuracy and its standard deviation, and the number of selected features for each method on the mnist database. The best results are highlighted in bold.

\begin{table}[htp]
	\footnotesize
	\centering
	\caption{Classification accuracy and the number of selected features.}\label{table:3}
	\begin{tabular}{llllll}
		\hline
		$\mu_{\min}$ &$\mu_{\max}$& KLR & $L_1$-norm RKLR & IKLR & $L_1$-norm RIKLR \\
		\hline
		-624.41 & 25689.67 & 0.509 $\pm$ 0.013 (396) & 0.514 $\pm$ 0.017 (129)&  0.966 $\pm$ 0.006 (396)& $\boldsymbol {0.968 \pm 0.008 (140)}$\\
		\hline

	\end{tabular}
\end{table}

It is worth noting that on this database, the TL1 kernel shows highly indefinite. For the sake of simplicity, we still compare the proposed $L_1$-norm RIKLR, KLR, $L_1$-norm RKLR and IKLR.
There is no significant difference in classification accuracy between $L_1$-norm RIKLR and IKLR in the indefinite kernel method, but $L_1$-norm RIKLR tends to choose a smaller number of features, demonstrating the effectiveness of the proposed model in achieving sparsity without compromising accuracy. Similarly, in the PD kernel method, the classification accuracy of $L_1$-norm RKLR is slightly higher than that of KLR, and $L_1$-norm RKLR selects a smaller number of features. On the whole, due to the high indefinite of the database, the performance of the indefinite kernel method is significantly better than that of the PD kernel method.

\section{Concluding remark}\label{sec:conclusion}

In this paper, we proposed the $L_1$-norm RIKLR model, which generalizes the IKLR model by incorporating the $L_1$-norm penalty. The nonsmooth property of the $L_1$-norm penalty function makes the existing optimization algorithm for the IKLR model unapplicable. To solve our nonsmooth and nonconvex optimization problem, we first formulated its objective function as DC functions, and then designed an efficient proximal linearized algorithm. The convergence theory of the proposed algorithm was also given. To illustrate the superiority of the proposed model, we conducted numerical experiments on various datasets and its comparison with other related KLR models. The extensive numerical results show that when the indefinite kernel captures much more important information the proposed $L_1$-norm RIKLR model will give a remarkable classification accuracy with much less features.

The $L_1$-norm penalty is a convex function, which not only makes our learning problem can be easily solved with existing R packages but also is important for decomposing the objective function as DC functions. It is also well known that many other penalty functions can induce different levels of sparse coefficient vector, and some of them may be nonconvex.
If we consider the nonconvex penalty function, the optimization procedure in the current paper may fail to give  satisfactory solution.
Thus, it should be of interest to consider other interesting penalty functions for the IKLR model and design its corresponding optimization algorithms. We leave it as the near future work.

\section*{Acknowledgement}\label{sec:acknowledgement}
The work was supported by National Natural Science Foundation of China (Grant No. 12301346) and Shandong Province Higher Education Youth Innovation and Technology Support Program (Grant No. 2023KJ199). The authors would like to thank Dr. Fanghui Liu for sharing the codes used in \cite{LHGYS18}.

\end{document}